\documentclass[letterpaper, 10 pt, conference]{ieeeconf}
\usepackage{cite}
\usepackage{amsmath,amssymb,amsfonts} 
\usepackage{graphicx}
\usepackage{textcomp} 
\usepackage{multirow}
\usepackage{gensymb}
\usepackage{xcolor} 
\usepackage[ruled,vlined]{algorithm2e}
\usepackage{gensymb}
\usepackage{textcomp} 
\usepackage{xcolor} 
\usepackage{color}
\usepackage{multirow}
\usepackage{graphicx}
\usepackage{subfigure}
\usepackage[mathscr]{eucal}
\usepackage{bm}
\usepackage[colorlinks,linkcolor=blue]{hyperref}
\usepackage{threeparttable}
\usepackage{caption}

\IEEEoverridecommandlockouts                              

\makeatletter

\makeatletter
\renewcommand{\maketag@@@}[1]{\hbox{\m@th\normalsize\normalfont#1}}%

\setbox0\hbox{$\xdef\scriptratio{\strip@pt\dimexpr
    \numexpr(\sf@size*65536)/\f@size sp}$}

\newcommand{\scriptveryshortarrow}[1][3pt]{{%
    \vcenter{\hbox{\rule[\scriptratio\dimexpr-.2pt\relax]
               {\scriptratio\dimexpr#1\relax}{\scriptratio\dimexpr.4pt\relax}}}%
   \mkern-4mu\hbox{\let\f@size\sf@size\usefont{U}{lasy}{m}{n}\symbol{41}}}}

\makeatother

\title{\LARGE \bf
SensorX2car: Sensors-to-car calibration for autonomous driving in road scenarios
}

\author{Guohang Yan$^{*}$, Zhaotong Luo$^{*}$, Zhuochun Liu and Yikang Li$^{\dagger}$ 
\thanks{$^{*}$ Equally contributed to the work.}
\thanks{$^{\dagger}$ Corresponding author.}
\thanks{Guohang Yan, Zhaotong Luo, Zhuochun Liu and Yikang Li are with Autonomous Driving Group, Shanghai Artificial Intelligence Laboratory, China. {\tt\small \{yanguohang, luozhaotong, liuzhuochun, liyikang\}@pjlab.org.cn}}
}

\begin{document}
\maketitle
\begin{abstract}
Properly-calibrated sensors are the prerequisite for a dependable autonomous driving system. However, most prior methods focus on extrinsic calibration between sensors, and few focus on the misalignment between the sensors and the vehicle coordinate system. Existing targetless approaches rely on specific prior knowledge, such as driving routes and road features, to handle this misalignment. This work removes these limitations and proposes more general calibration methods for four commonly used sensors: Camera, LiDAR, GNSS/INS, and millimeter-wave Radar. By utilizing sensor-specific patterns: image feature, 3D LiDAR points, GNSS/INS solved pose, and radar speed, we design four corresponding methods to mainly calibrate the rotation from sensor to car during normal driving within minutes, composing a toolbox named SensorX2car. Real-world and simulated experiments demonstrate the practicality of our proposed methods. Meanwhile, the related codes have been open-sourced to benefit the community. To the best of our knowledge, SensorX2car is the first open-source sensor-to-car calibration toolbox. The code is available at \href{https://github.com/OpenCalib/SensorX2car}{https://github.com/OpenCalib/SensorX2car}.

\end{abstract}

\section{INTRODUCTION}
Autonomous driving vehicles perceive the environment through sensors. The performance of sensors is limited by the quality of sensor calibration \cite{dang2009continuous}. The current prominent solution relies on accurate sensor installation in the factory in the first place or performs offline calibration in artificially-arranged rooms which contain specially-designed facilities \cite{alberts2019robust,koppanyi2018experiences,persic2017extrinsic} and markers \cite{zhang2000flexible, meng2003new}. Although these methods can provide adequate precision, they are time-consuming and require expensive equipment, making it hard to realize in users' hands. Unfortunately, due to the vibration and operating condition changes in daily use, the extrinsic parameter of sensors won't stay the same. This calls for online target-free calibration methods. 

\begin{figure}
    \centering
    \includegraphics[width=8.5cm]{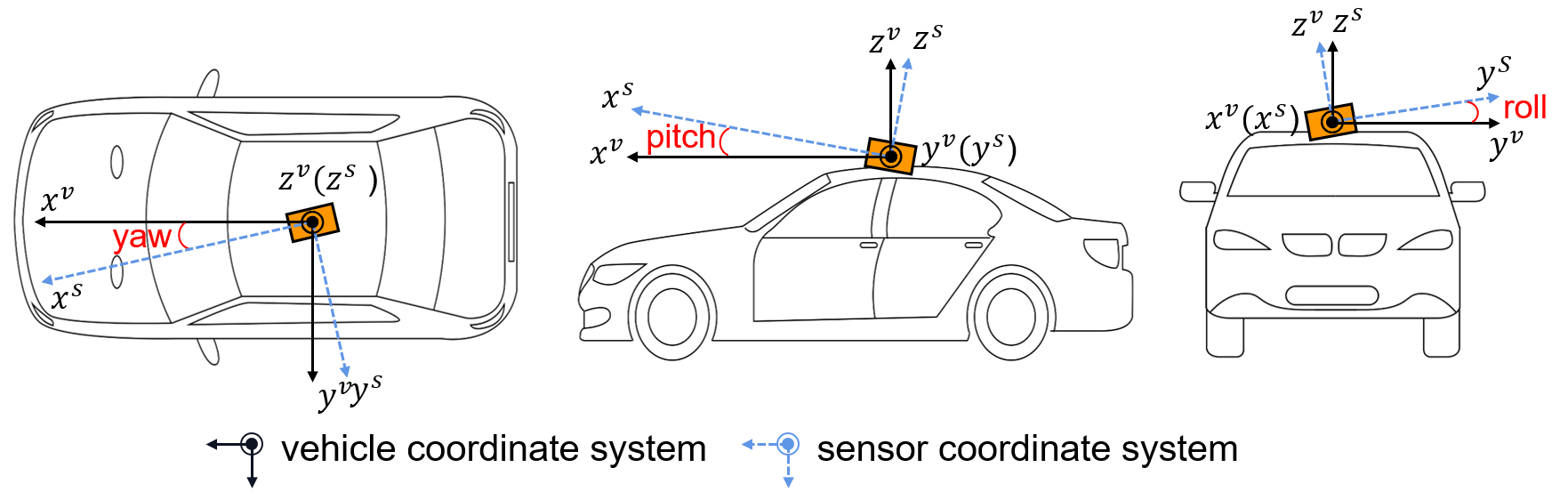}
    \caption{Schematic diagram of Sensor-to-car calibration. We only calibrate the three angles (yaw, pitch, roll) from the sensor coordinate system to the vehicle coordinate system.}
    \label{fig:coordinate}
\end{figure}

Currently, most online methods are concerned with calibration between sensors. Focused topics are camera-LiDAR systems \cite{li2022automatic}, multi-camera systems \cite{ince2020accurate, dexheimer2022information} and visual-inertial systems \cite{huang2018online, liu2019visual, chu2020keyframe}. Few studies deal with the car-body frame. However, the extrinsic between the sensor and vehicle, especially the rotation, is also essential in many autonomous driving tasks, such as automated braking and lane keeping. Sensors must be well aligned with the car orientation as expected to ensure accurate decision making. Thus, we present SensorX2car, containing online calibration methods for four commonly-used sensors: camera, LiDAR (Light Detection and Ranging), GNSS/INS (Inertial Navigation System) device, and 2D millimeter wave radar. The calibrated Euler angles from the sensor to the car-body frame are shown in Fig. \ref{fig:coordinate}. Note that we move the origins of the two coordinate systems together for clear illustration, which may not be the actual case.

We designed four corresponding methods according to the characteristics of each sensor. For the camera, we use a deep learning network to predict the vanishing point and horizon line of a single image and then convert them to the roll, pitch, and yaw angles. For LiDAR, we use simultaneous localization and mapping (SLAM) algorithm to generate the 6 DoF pose. The yaw angle is calibrated by averaging the difference between the vehicle forward direction and LiDAR forward direction. The roll and pitch angle is calibrated by ground extraction. For GNSS/INS integrated device, we calibrate the yaw angle using the 6 DoF pose it outputs. For millimeter wave radar, we utilize the principle of radar speed and calibrate yaw by fitting a cosine curve. 

In these methods, the acquired rotation is mainly between the sensor and the world coordinate system. In order to get the rotation from the vehicle to the world frame, we use some assumptions. Ordinarily, the vehicle's orientation is consistent with its trajectory, velocity, or road direction. These assumptions inevitably introduce errors, so in the experimental practice, we accumulate data for a period of time and filter out some unreliable points to get a final result. 

To our knowledge, there are few methods and open-source codes for sensor-to-car calibration \cite{heide2018calibration,meyer2021automatic,rodrigo2020novel,kellner2015joint}. The existing ones are conditioned by equipment, road features, or driving route. Our methods basically removed these limitations by not relying on specific road features like lane lines, and make it theoretically capable of handling arbitrary driving routes, though it will work better on a straight road. This work is based on the previous OpenCalib \cite{yan2022opencalib} research and project experience. The related code has been open-sourced on GitHub to benefit the community.

The contributions of this work is listed as follows: 
\begin{enumerate}
\item We propose a camera-to-car calibration method by using a deep learning network, which can calibrate the three rotation angles without relying on lane lines. We build a dataset based on KITTI, train and test on it.
\item We propose a LiDAR-to-car calibration method by using vehicle trajectory and ground extraction, which can calibrate the three rotation angles and the height to the ground without restrictions on the driving routes.
\item We propose a GNSS/INS-to-car calibration method by using vehicle trajectory to calibrate the heading (yaw) angle.
\item We propose a radar-to-car calibration method by using detected object speed and vehicle speed, which can calibrate the heading (yaw) angle. We optimize this method by adding position information to solve the large noise problem in the real-world dataset.
\item The methods are verified by simulation and real-world data. The related code is open-sourced to benefit the community.
\end{enumerate}

\section{RELATED WORK} 
In factory or laboratory, the extrinsic of sensor-to-car can be calibrated by direct measuring and using target with specific patterns or geometries. For camera, the fundamental method is using man-made markers like chessboard to minimize re-projection errors (RPEs) in image space \cite{zhang2000flexible}. For LiDAR, common approaches use unique geometric target like cylinder \cite{alberts2019robust} or reflective target placed on a wall \cite{koppanyi2018experiences}. For radar, it is usually calibrated by trihedral corner reflector \cite{persic2017extrinsic}. Despite calibrating with these specific facilities, we then discuss some online methods as follows.

\subsection{Camera-to-car Calibration}

For online camera calibration, the mainstream method is detecting vanishing points(VPs), which is the intersection of image lines from the 2D perspective projections of mutually parallel lines in 3D space. It can generate the camera pose in the world coordinate system. Conventional VP detection depends on line features \cite{zhou2017detecting}. The overall process can be divided into two stages: line extraction and line clustering.
However, these methods cannot produce a holistic understanding of the scene. Learning-based method handles this problem by formulating it as a regression or classification problem. \cite{chang2018deepvp} \cite{zhang2018dominant} divide the image into multiple grids and classify each grid according to VP position. \cite{lee2021ctrl} \cite{kluger2017deep} regress the sphere image representation of vanishing point. \cite{honda2021end} proposed an end-to-end network to predict a VP heatmap. Geometric priors of vanishing point are exploited to improve the robustness and accuracy of the algorithm and reduce data reliance \cite{kluger2017deep}\cite{zhou2019neurvps}\cite{lin2022deep}. 


However, it needs more than one VP to calibrate the rotation. In many driving scenes, there only exist one dominant VP, which is the VP of the road. This can only give us 2 DoF of the rotation. In order to complete the other degree, we additionally estimate the horizon line, which is the intersection of the parallel of any original plane and the picture. Current horizon estimation methods mainly rely on supervised learning and deep networks \cite{workman2016horizon} \cite{kluger2020temporally}.


\subsection{LiDAR-to-car Calibration}
Most online calibration methods about LiDAR have been focused on multi-LiDAR system or LiDAR fused with other different type of sensor, especially camera. 

Fewer methods concerning about calibration between LiDAR and car-body coordinate system. \cite{heide2018calibration} calibrate the extrinsic by registering LiDAR points to a 3D model of the vehicle as the prior knowledge. Meyer et al. \cite{meyer2021automatic} proposed a method to calibrate yaw by detecting edge or lane in road, and calibrate pitch and roll by detecting the planar orientation of the ground surface just in front of the vehicle. It requires that the vehicle travels along a straight line with a well-marked road edge. Our method doesn't have such requirements.

\subsection{GNSS/INS Heading Calibration}

GNSS-aided INS system is a common combination to provide better dynamic performance. Previous work \cite{niu2008improving} has pointed out the importance of knowing the IMU mounting angles while using both IMU and odometer or Non-holonomic Constraints (NHC) in navigation. Vinande et al. \cite{vinande2009mounting} estimates the mounting misalignment of an IMU using forward and rearward acceleration and vehicle orientation from GPS, but the estimation error is relatively large to 2$\degree$. Bao et al. \cite{bao2013calibration} calibrate the misalignment angle of vehicle-mounted IMU by measuring the specific force when accelerating or braking frequently, assuming the vehicle is travelling on a horizontal plane and making a straight motion. Without GNSS, Rodrigo et al.\cite{rodrigo2020novel} formulates a joint state problem and solve it by an adaptive Kalman filter, using vehicle motion information from the odometry and suspension systems. However, it needs aid from artificial measurements in a standstill phase. Compared, our method performs online calibration by using the output of GNSS/INS integrated devices without other dependencies. We only calibrate the heading(yaw) angle, which is the most important.

\subsection{Radar-to-car Calibration}

For 2D millimeter wave radar, we only need to calibrate the heading (yaw) angle. Izquierdo et al. \cite{Izquierdo2018} proposed an automatic and unsupervised method by recording highly sensitive static elements like lights or traffic signs in a high-definition (HD) map, then obtaining the extrinsic calibration by registering radar scans to the corresponding targets. However, this method requires a very precise localization (less than 2 cm) in the HD map. \cite{kellner2015joint} analyze the Doppler distribution of stationary targets over the azimuth angle while the vehicle is driving on an approximately straight line. Our method can work on curve driving route without other dependencies.

\section{METHODOLOGY}

\subsection{Camera-to-car Calibration}

\begin{figure}[htp]
    \centering
    \includegraphics[width=\linewidth]{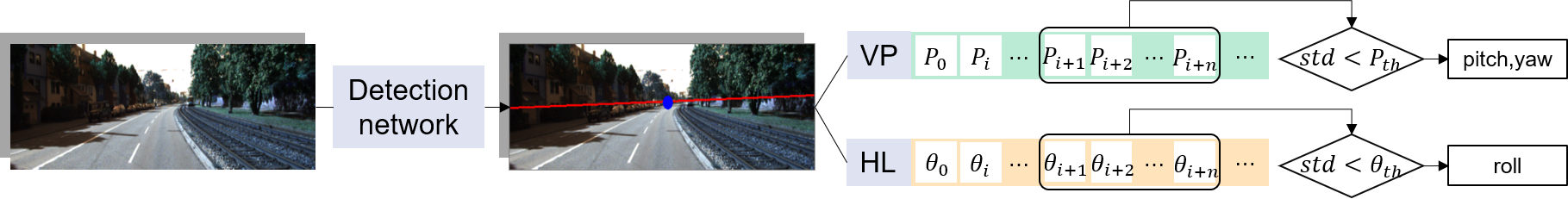}
    \caption{Camera calibration process: Input an image, the detection network outputs the vanishing point and horizon line angle. If the fluctuation of a fixed number of VPs and angles is small enough, they are used to calibrate the angles.}
    \label{fig:camera_process}
\end{figure}

The overall process for camera calibration is shown in Fig. \ref{fig:camera_process}. We first estimate the vanishing point (VP) and the horizon line (HL) angles of every image by a deep learning network. Assuming the intrinsic of the camera is already known, we can calculate the rotation from the camera to the road. Ideally, the XY plane of the car-body is parallel to the ground, and the heading direction of the vehicle is the same as the road direction while driving straightly. Thus, we equivalently get the rotation from the camera to the car. In order to handle the vehicle turning and uneven roads, we use the stability of the output over a period of time as a judgment condition to extract useful driving segments.

\subsubsection{Detection Network}
\

\begin{figure}[htp]
    \centering
    \includegraphics[width=\linewidth]{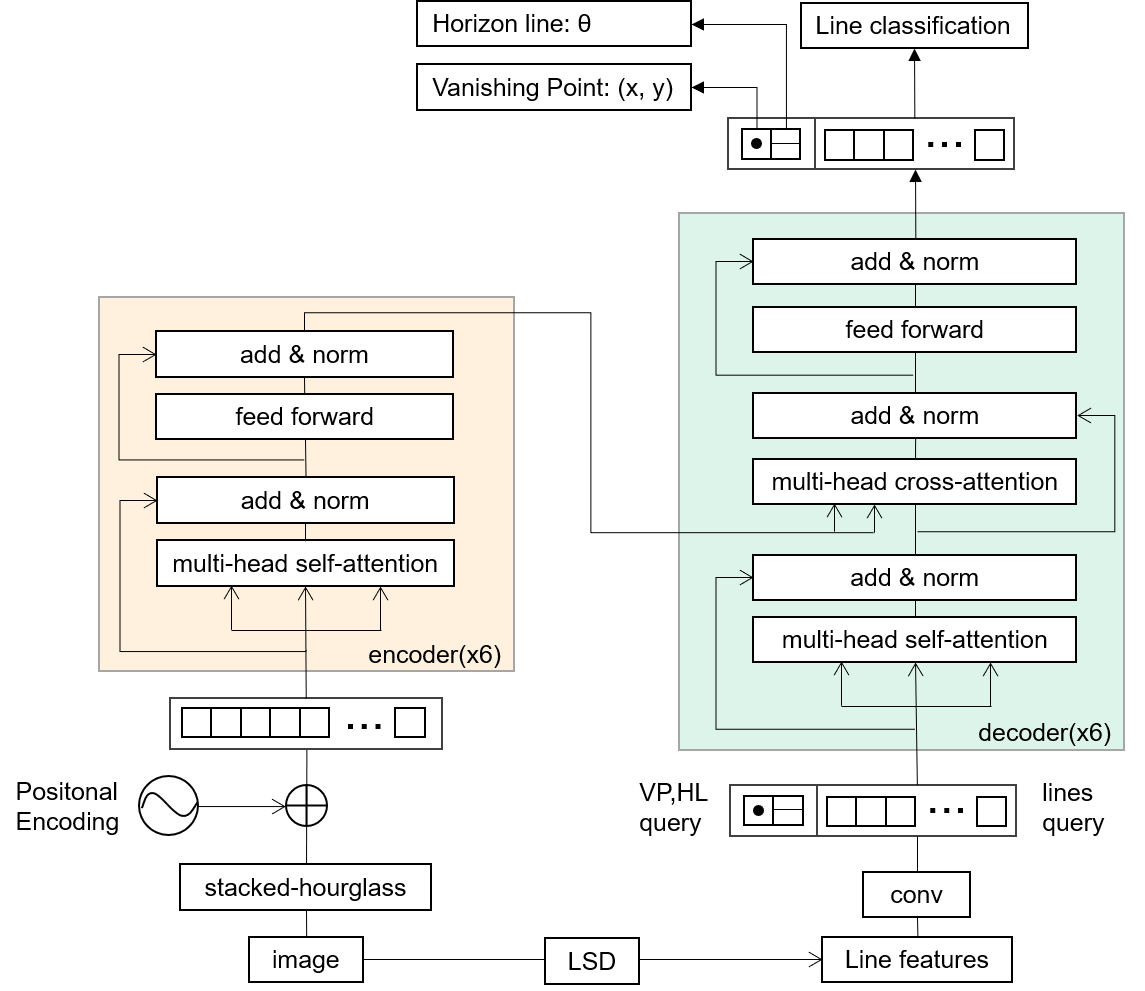}
    \caption{The overview of the network. From an input image, the feature map is generated by the stacked-hourglass backbone and line features are extracted by LSD. The outputs are VP position, HL angle and line feature labels to calculate the overall loss.}
    \label{fig:network}
\end{figure}

The overview of the detection network is shown in Fig. \ref{fig:network}. The network estimates vanishing point position and horizon line angle. Referring to \cite{lee2021ctrl}, we use Transformer as the basic structure and feed the extracted line features to decoder layers.

We use Stacked-Hourglass network\cite{newell2016stacked} as the backbone to generate the feature map, as it can remain local details while expanding the receptive field. 
For an input image $( 3, H_0, W_0)$, the high-dimensional feature map $( C, H, W)$ from the backbone is flattened to $( C, H \times W )$ and fed into the encoder layers. 

Two learned queries are set as the input of the decoder layers to estimate VP and HL. Besides, the extracted lines also form queries after a convolution network. At the output end of the decoder, a classification problem is formulated to distinguish whether a line passes through the vanishing point. In this way, the network is encouraged to pay more attention to the lines passing through the vanishing point and have better performance. 

The distance between a line $\textbf{l}_i$ and the vanishing point $\textbf{p}_{gt}$ can be represented as:
\begin{equation}
    d_i = \left| \frac{\textbf{l}_i \cdot \textbf{p}_{gt}}{\left \| \textbf{l}_i \right \| \left \| \textbf{p}_{gt} \right \|}\right|
\end{equation}
We set a threshold $d_{th}$ to distinguish positive and negative samples.
\begin{equation}
    c_i = \left\{ 
    \begin{matrix}
    1 \qquad d_i < d_{th} \\
    0 \qquad d_i > d_{th}
    \end{matrix}
    \right.
\end{equation}

The ultimate loss function consists three components: the cosine similarity of the vanishing point position, the absolute value error of the horizon line angle, and the BCE (Binary Cross Entropy) loss of line classification. 

\begin{equation}
    loss = l_{V} + l_{H} + l_{L}
\end{equation}

\begin{equation}
    l_{V} = 1 - \left| \frac{\textbf{p}_{gt} \cdot \textbf{p}}{\left \| \textbf{p}_{gt} \right \| \left \| \textbf{p} \right \|}\right|
\end{equation}

\begin{equation}
    l_{H} = \left| \theta - \theta_{gt} \right|
\end{equation}

\begin{equation}
    l_{l} = -\frac{1}{n} \sum \left \{ c_i log(s_i) + (1-c_i) log(1 - s_i) \right \}
\end{equation}

\subsubsection{Rotation Derivation}
\

In identify useful driving periods, we continuously calculate the standard deviation (Std) of a fixed number of VPs and angles. If the Std falls below a threshold, we consider as the car is driving on a straight road and the road is flat, where we output the average rotation by Eq.\ref{eq:camera_R3}-\ref{eq:camera_roll} as the calibration result. 

Using vanishing point for camera calibration has been developed for decades. Here we just present the final formulation from VP to angles. The derivation details can be found in \cite{caprile1990using}.

\begin{equation}
    R_3 =  \frac{K^{-1}p_\infty}{||K^{-1}p_\infty||} = \left [ \begin{matrix}
        r_1 \\ r_2 \\ r_3
    \end{matrix} \right ]
    \label{eq:camera_R3}
\end{equation}

\begin{equation}
    yaw = arcsin(r_2), \ pitch = -arctan( \frac{r_1}{r_3} )
\end{equation}


where $p_\infty$ is the vanishing point in the image plane. $R_3$ is the third column of the rotation matrix R. K is the intrinsic matrix.
The roll angle is just the angle of the horizon line.
\begin{equation}
    roll = \theta_{HL}
    \label{eq:camera_roll}
\end{equation}

\subsection{LiDAR-to-car Calibration}
For LiDAR, we calibrate its rotation to the car and height to the ground. In detail, we first run the SLAM algorithm to get the 6 DoF pose denoted as $\widetilde{T}_i$, which will be used in the following two procedures. In the experiment, we choose FAST-LOAM \cite{wang2021f} as it can achieve both real-time performance and good accuracy. For pitch and roll angle, we extract the ground plane from LiDAR points. For yaw angle, we analyze the difference between the vehicle heading direction and LiDAR forward direction at corresponding timestamps. 

\subsubsection{Pitch and Roll Estimation}
\ 

\begin{figure}[htp]
    \centering
    \includegraphics[width=\linewidth]{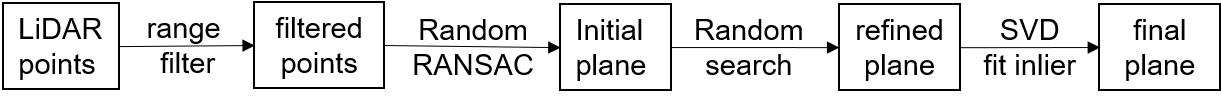}
    \caption{Ground extraction process. The LiDAR points are first filtered by distance, and then fitted by RANSAC to get an initial guess. The plane parameter is refined by random search, and the inliers are refitted to get the final result. }
    \label{fig:lidar_process}
\end{figure}

The procedure for ground extraction is shown in Fig. \ref{fig:lidar_process}. For an input LiDAR frame, we first filter out points too far or close. Then we apply the RANSAC algorithm multiple times with random initial points to get a coarse estimation. We refine this estimation by conducting a random search around it and adjusting the plane parameter in a small range to get a plane with larger inliers. The inliers are then fitted to the final plane using SVD.

The extracted plane can be represented by its normal vector $\vec{n}_p$ and intercept $d_p$:
\begin{equation}
    \begin{matrix}
    ax+by+cz+d = 0 \\
    \vec{n}_p = [a,\ b,\ c],\ d_p = d
    \end{matrix}
\end{equation}

The rotation vector $\vec{n}$ and rotation angle $\alpha$ from LiDAR to the road can be calculated as:
\begin{equation}
    \vec{n} = \vec{n}_p \times \vec{z}, \ \alpha = arcos(\vec{n}_p \cdot \vec{z})
\end{equation}

The rotation matrix R and height z is derived as:
\begin{equation}
    \begin{matrix}
    R = cos\alpha I +(1-cos\alpha) \vec{n}\vec{n}^T +sin\alpha n^\wedge \\
    z = \frac{d_p}{c}
    \end{matrix}
\end{equation}
where $\vec{z} = [0,\ 0,\ 1]^T$.

The above process is only performed on a downsampled set of LiDAR frames over time for real-time performance to avoid repeated calculations. We also filter out frames in sharp turns according to the pose from the SLAM algorithm because turning leads to vehicle tilt and error in roll angle. 

\subsubsection{Yaw Estimation}
\ 

After getting pitch and roll, we apply the rotation to the point cloud so that the z-axis of the point cloud is perpendicular to the ground plane.  

The heading direction of the vehicle is obtained by the tangent direction of the 2D trajectory curve on the XY plane. We apply B-spline interpolation on the timely-discrete position $(\widetilde{x}_{t=t_i} = \widetilde{x}_i, \widetilde{y}_{t=t_i} = \widetilde{y}_i)$ from SLAM algorithm:

\begin{equation}
    \begin{matrix}
    x(t) = Bspline(\{ \widetilde{x}_i \}, \{ t_i \}, n) \\
    y(t) = Bspline(\{ \widetilde{y}_i \}, \{ t_i \}, n) \\
    \end{matrix}
\end{equation}
where n is the degree of B-spline.

The heading direction of vehicle can be calculated by the derivative of $x(t)$ and $y(t)$ as:
\begin{equation}
    \begin{matrix}
    \psi_i^v = arctan(\frac{y'_i}{x'_i}) \\
    x'_i = \frac{dx(t)}{dt} \|_{t = t_i}, \ y'_i = \frac{dy(t)}{dt} \|_{t = t_i} \\
    \end{matrix}
\end{equation}

The forward direction of LiDAR can simply be obtained by the LiDAR pose, denoted as $\widetilde{\psi}_i$. So the difference between these two angles at the corresponding timestamp is the yaw offset that we need to calibrate. Besides, we filter out data points when the vehicle is driving at a low speed or navigating a highly curved route, to prevent erroneous direction estimation or poor curve fitness. The final result is the average estimation of all valid timestamps $S$:

\begin{equation}
    \psi_{offset} = \frac{1}{\left \| S\right \|} \sum_{i \in S}(\widetilde{\psi}_i - \psi_i^v)
    \label{eq:lidar_angle}
\end{equation}

The velocity and the curvature is calculated as: 
\begin{equation}
        v_i = (x'_i)^2 + (y'_i)^2
\end{equation}

\begin{equation}
        c^x_i = \frac{||x''_i||}{(1 + (x'_i)^2)^\frac{3}{2}}, \
        c^y_i = \frac{||y''_i||}{(1 + (y'_i)^2)^\frac{3}{2}}
\end{equation}


\subsection{GNSS/INS Heading Calibration}
For the GNSS/INS integrated device, we estimate the yaw angle of IMU and the vehicle to correct the installation error of forwarding orientation. 

GNSS/INS devices can provide 6 DoF pose with high accuracy and frequency. So it can follow the same way as LiDAR to calibrate the yaw angle. The formula is as follows:
\begin{equation}
    \psi_{offset} = \frac{1}{\left \| S \right \|} \sum_{i \in S}(\psi_i^I - \psi_i^v)
    \label{eq:IMU_angle}
\end{equation}
where $\psi_i^v$ is the vehicle direction derived from B-spline interpolation, $\psi_i^I$ is the IMU direction in the output pose, S is the valid time set selected by velocity and curvature.


\subsection{Radar heading Calibration}

We calibrate the heading misalignment of radar and the vehicle by the relationship between the detected object speed and vehicle speed. Our primary method only needs vehicle speed obtained from other sensors, such as a speedometer. 

\begin{figure}[tp]
    \centering
    \includegraphics[width=8cm]{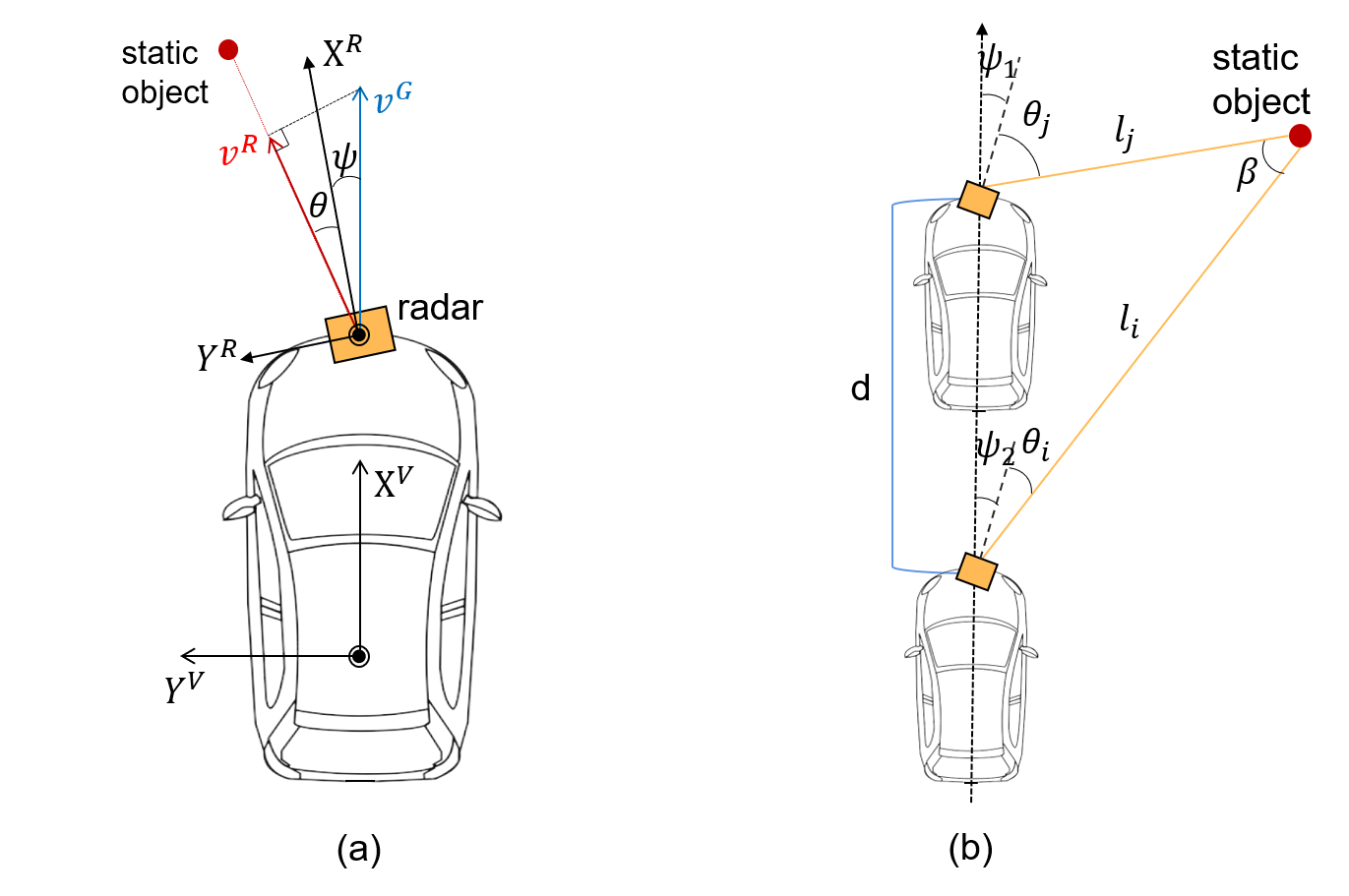}
    \caption{(a) For static objects, the velocity $v^R$ detected by radar is the component of vehicle velocity $v^G$ in the object direction. (b) The geometric relationship between vehicle position, radar measurements and radar misalignment angle.}
    \label{fig:radar}
\end{figure}

The radar measurement at $t_i$ can be denoted as $\{v^R_i, \theta^R_i, l^R_i, s_i\}$, which is the relative velocity, azimuth angle, distance, and track ID of a detected object. The vehicle speed is $v_i^G$ at $t_i$. As shown in Fig. \ref{fig:radar}(a), if the object is stationary, the detected velocity $v^R_i$ is equal to the component of vehicle speed $v_i^G$ in the object direction. The offset yaw angle $\psi$ satisfies the equation:
\begin{equation}
    v^R_i = v_i^G cos(\theta^R_i + \psi) \label{eq:radar_vel}
\end{equation}

By applying this basic formula, the calibration procedure can be divided into two steps: 

\subsubsection{Course Calibration}
In a certain range $[-A, A]$, the yaw candidates are uniformly sampled with interval n: $\psi_i = -A + i * n$. Then we test these candidates with a subset of radar points by Eq.\ref{eq:radar_vel}. The yaw angle with the most points satisfying this condition is used as the initial estimate. Here the assumption is that the static objects account for the majority.

\subsubsection{Refined Calibration}
The following procedures are done iteratively to each frame: with an initial guess of yaw, and we first pick out the stationary points by Eq.\ref{eq:radar_vel}; then we use these points to fit the cos function curve and update the yaw estimation. After a certain number of iterations, the estimates are recorded and averaged for a final result.

However, due to the large noise in radar measurement in practice, we improved this method by informing the position of the vehicle $\{x^G_i, y^R_i\}$, which can help build stronger constraints, as shown in Fig. \ref{fig:radar}. It is done as follows:

\subsubsection*{1) Object grouping}
We first extract straight-line segments from driving routes. Turning trajectories can be approximated as polylines, and the segments of sufficient length are preserved. In these segments, the radar points are grouped into objects by their track ID. If the same ID is uninterruptedly tracked in a number of frames, we consider the points with this ID as one object having the same position. These measurements are grouped into one object $O_n = \{ P_i = \{v^R_i, \theta^R_i, l^R_i, s_i, v^G_i, x^G_i, y^G_i \}  | s_i = s\}$.

\subsubsection*{2) Static object calibration}
For an object $O_n$, it is considered stationary if all point pairs $P_i(i=1,2,...)$ and $P_0$ satisfy:
\begin{equation}
    |arcos(\frac{l^2_i+l^2_0-d^2}{2 l_i l_0}) - (\theta_i - \theta_0)| < e
    \label{eq:radar_angle_1}
\end{equation}

The stationary objects are used to estimate the yaw angle. For point $P_i(i=1,2,...)$, the angle $\psi_i$ is calculated by:
\begin{equation}
    \begin{matrix}
        \psi_1 = arcos(\frac{l^2_i+d^2-l^2_0}{2 d l_i }) - \theta_i \\
        \psi_2 = arcos(\frac{l^2_0+d^2-l^2_i}{2 d l_0 }) - \theta_0 \\
        \psi_i = (\psi_1 + \psi_2)/2
    \end{matrix}
    \label{eq:radar_angle_2}
\end{equation}

The confidence of the estimation $\psi$ is calculated by the velocity:

\begin{equation}
    c_i = 1 - \frac{|v_i^R/cos(\theta_i^R+\psi_i)-v_i^G|}{v_i^G}
\end{equation}

The final result is the weighted average of estimations from all stationary objects.
\begin{equation}
    \psi = \sum_{O_n\in O} \sum_{P_i\in O_n} c_i\psi_i
\end{equation}

\section{EXPERIMENTS}
\vspace{-1mm}

In this part, we perform qualitative and quantitative experiments to prove the accuracy and adaptability of our proposed methods. However, one obstacle in calibration validation is the lack of ground-truth reference against which the estimates can be directly compared. Since the extrinsic won't change in a short period of time, for real-world data, we use the stability of the estimation as a metric. In addition, we also visualized part of the calibration process of each method to verify them further.

\subsection{Camera Calibration}

\subsubsection{Dataset}
\begin{figure}[htp]
    \centering
    \includegraphics[width=\linewidth]{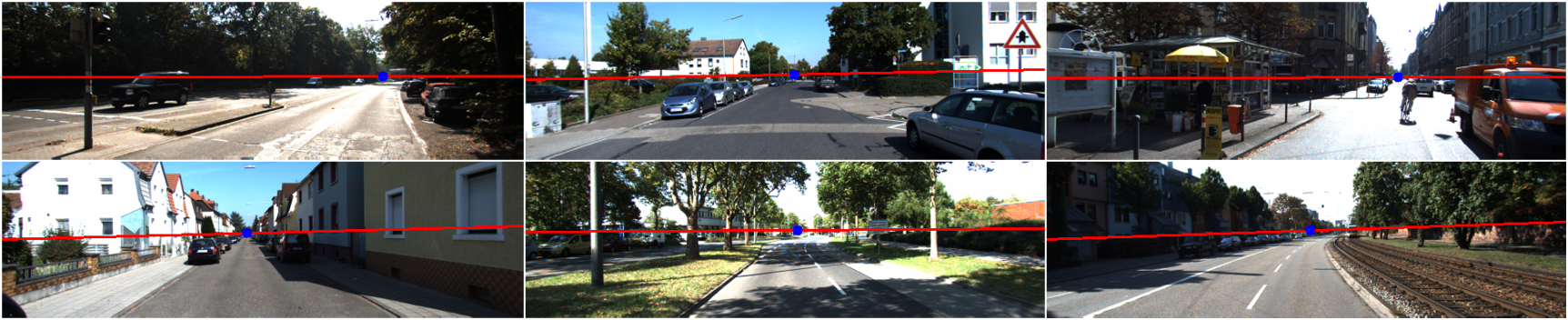}
    \caption{Examples in annotated KITTI dataset, including city, residential, highway, campus, mountainous scenes etc.}
    \label{fig:kitti}
\end{figure}

In order to train and evaluate our detection network, we need a dataset of sufficient size. However, we find few open-source datasets contain both vanishing point and horizon line annotations while also being suitable for autonomous driving scenarios. So we annotate our dataset based on KITTI\cite{geiger2013vision}. The horizon line is derived automatically by the absolute pose of IMU and the relative pose between IMU and the camera, referred to \cite{kluger2020temporally}. The vanishing point is first predicted by another network Neurvps\cite{zhou2019neurvps} and then corrected manually. The final dataset contains 43195 images with different scenes. Fig.\ref{fig:kitti} shows some examples. 

\subsubsection{Network}
Since our main target is calibration, we compare our method simply with two other well-performing networks. One for vanishing point detection \cite{zhou2019neurvps} and one for horizontal line estimation \cite{kluger2020temporally}. They were both trained and tested on the annotated KITTI dataset, and the result is shown in Tab. \ref{tab:network_result}. It demonstrated that our method has competitive performance and a relatively high inference speed while achieving both two tasks. 

\begin{table}[htp]
    \centering
    \caption{Quantitative evaluation result of the network}
    \begin{tabular}{c|c|c|c|c}
         \hline
         \multirow{2}*{Model} & VP(pixel) & \multicolumn{2}{|c|}{HL(rad)} & \multirow{2}*{FPS} \\
         \cline{2-4}
         ~ & L2 distance & $MSE\times10^{-3}$ & AUC & ~ \\
         \hline
         Neurvps\cite{zhou2019neurvps} & 10.4623 & $\backslash$ & $\backslash$ & 1.138 \\
         \hline
         TCHL\cite{kluger2020temporally} & $\backslash$ & 0.298 & 75.37$\%$ & 20.2432 \\
         \hline
         ours & 7.627 & 0.348 & 71.6$\%$ & 22.6482 \\
         \hline
    \end{tabular}
    \label{tab:network_result}
\end{table}

\subsubsection{Calibration result}
KITTI doesn't offer the reference value of camera-to-car extrinsic. In order to validate the feasibility of our method, we conduct the experiment in two ways.

\begin{figure}[htp]
    \centering
    \includegraphics[width=\linewidth]{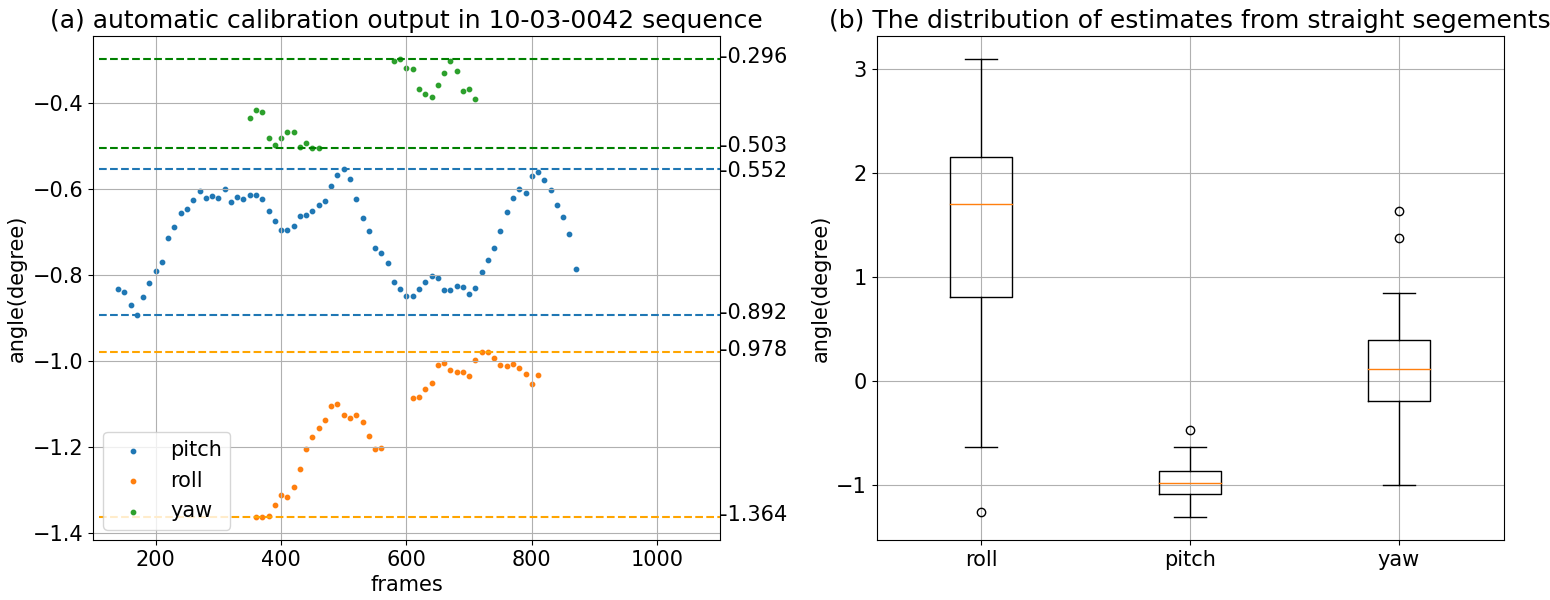}
    \caption{(a) The automatic calibration result in 2011-10-03-42 sequence. The fluctuation range of the three angles is about $0.2\degree-0.4\degree$. (b) The distribution of calibration estimates in 72 straight driving sequences. The fluctuation range of roll, pitch, yaw is about 1.1$\degree$, 0.2$\degree$, 0.5$\degree$}
    \label{fig:camera_result}
\end{figure}

The algorithm first runs automatically on a complete piece of data, covering turns and slopes. 
The algorithm output calibration values successively when the standard deviation of estimations from a fix number (n=100) of frames falls below a threshold (e = 0.005 rad). The output in is shown in Fig. \ref{fig:camera_result}(a). This sequences lasts about 1000 frames. The fluctuation range of the three angles is smaller than 0.4$\degree$.

Since the first experiment limits the fluctuation range of the angle, the consistency of estimations loses the verification meaning, so we conducted another experiment. We selected 72 sequences $\{S_1,S_2,...,S_{72}\}$ from KITTI when the car is driving continuously in a straight line and on a flat road. Each contains 100-300 images. Then we get a final estimation $\{r_i,p_i,y_i\}$ for each sequence by averaging the outputs of all images in $S_i$. The distribution of the 72 estimates is shown in Fig. \ref{fig:camera_result}(b). The standard deviation of estimations in different dates is shown in Tab. \ref{tab:camera_result}. Here we suppose the extrinsic doesn't change much through the whole dataset. The fluctuation range of pitch and yaw angle is around 0.2$\degree$ and 0.5$\degree$, respectively. The roll angle has a relatively large fluctuation, probably because the roll angle between the vehicle and the ground changes greatly when driving.

\begin{table}[htp]
    \centering
    \caption{Calibration consistency for straight-line sequences}
    \begin{tabular}{c|c|c|c|c}
         \hline
         date & seqs num & roll(deg) & pitch(deg) & yaw(deg) \\
         \hline
         all date & 72 & 1.112597 & 0.170348 & 0.4743682 \\
         \hline
         2011-09-26 & 28 & 0.927527 & 0.167576 & 0.396274 \\
         \hline
         2011-09-30 & 27 & 1.072881 & 0.134185 & 0.455136 \\
         \hline
         2011-10-03 & 15 & 1.138978 & 0.121006 & 0.576611\\
         \hline
    \end{tabular}
    \label{tab:camera_result}
\end{table}

\begin{table*}
    \vspace{0.2cm}
    \caption{Calibration result of four periods of real-world data for LiDAR, GNSS/INS and Radar (all/straight)}
    \centering
    \begin{tabular}{c|c|c|c|c|c|c|c|c|c|c}
        \hline
        \multirow{2}*{data} & seqs & \multicolumn{5}{c|}{LiDAR} & \multicolumn{2}{c|}{GNSS/INS} & \multicolumn{2}{c}{Radar}\\
        \cline{3-11}
        ~ & num & roll Std(deg) & pitch Std(deg) & yaw Std(deg) & z Std(m) & time & yaw Std(deg) & time & yaw Std(deg) & time \\
        \hline
        1 & 26/14 & 0.5135/0.1414 & 0.0886/0.0869 & 0.0646/0.0266 & 0.0676/0.0471 & 98.2 & 0.1034/0.1241 & 0.311 & 0.8589/0.3032 & 0.828\\
        \hline
        2 & 32/12 & 0.6880/0.1284 & 0.1051/0.1160 & 0.0474/0.0480 & 0.0689/0.0467 & 95.7 & 0.2427/0.1353 & 0.321 & 0.3986/0.3358 & 0.792\\
        \hline
        3 & 28/14 & 0.4115/0.1132 & 0.0570/0.0429 & 0.1461/0.1014 & 0.0550/0.0354 & 92.6 & 0.0659/0.0550 & 0.390 & 0.7673/0.4001 & 0.732\\
        \hline
        4 & 28/12 & 0.6705/0.1224 & 0.1164/0.1432 & 0.0430/0.0380 & 0.0572/0.0649 & 96.4 & 0.0440/0.0278 & 0.297 & 0.3909/0.3371 & 0.838\\
        \hline
    \end{tabular}
    \begin{tablenotes}
    \item all/straight: The preceding number is the result for all segments, and the following number is the result for only straight-driving segments. 
    \item time: the average running time(seconds) of each algorithm for a 1-minute data segment
    \end{tablenotes}
    \label{tab:lidar_result}
\end{table*} 

\subsection{LiDAR Calibration}
\subsubsection{Dataset}
\begin{figure}[tp]
    \vspace{0.2cm}
    \centering
    \includegraphics[width=\linewidth]{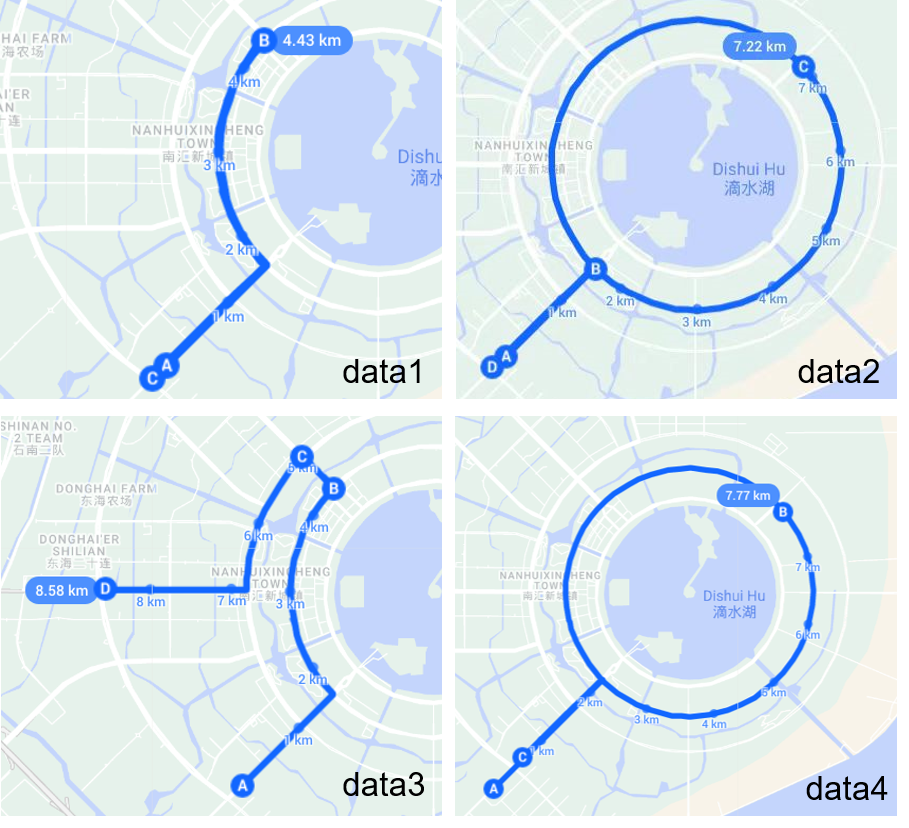}
    \caption{Driving route of four pieces of data collected on different dates. data1: A-B-C(9km), data2: A-B-C-D(14.5km), data3: A-B-C-D(8.5km), data4: A-B-C(15km). }
    \label{fig:map}
\end{figure}

We test our method on four periods of real-world data, each lasting around thirty minutes. The data is collected by a Pandar64 LiDAR mounted on the top of the vehicle. The driving route of four segments of data is shown in Fig. \ref{fig:map} by Google Maps.


\begin{figure}[htp]
    \centering
    \includegraphics[width=\linewidth]{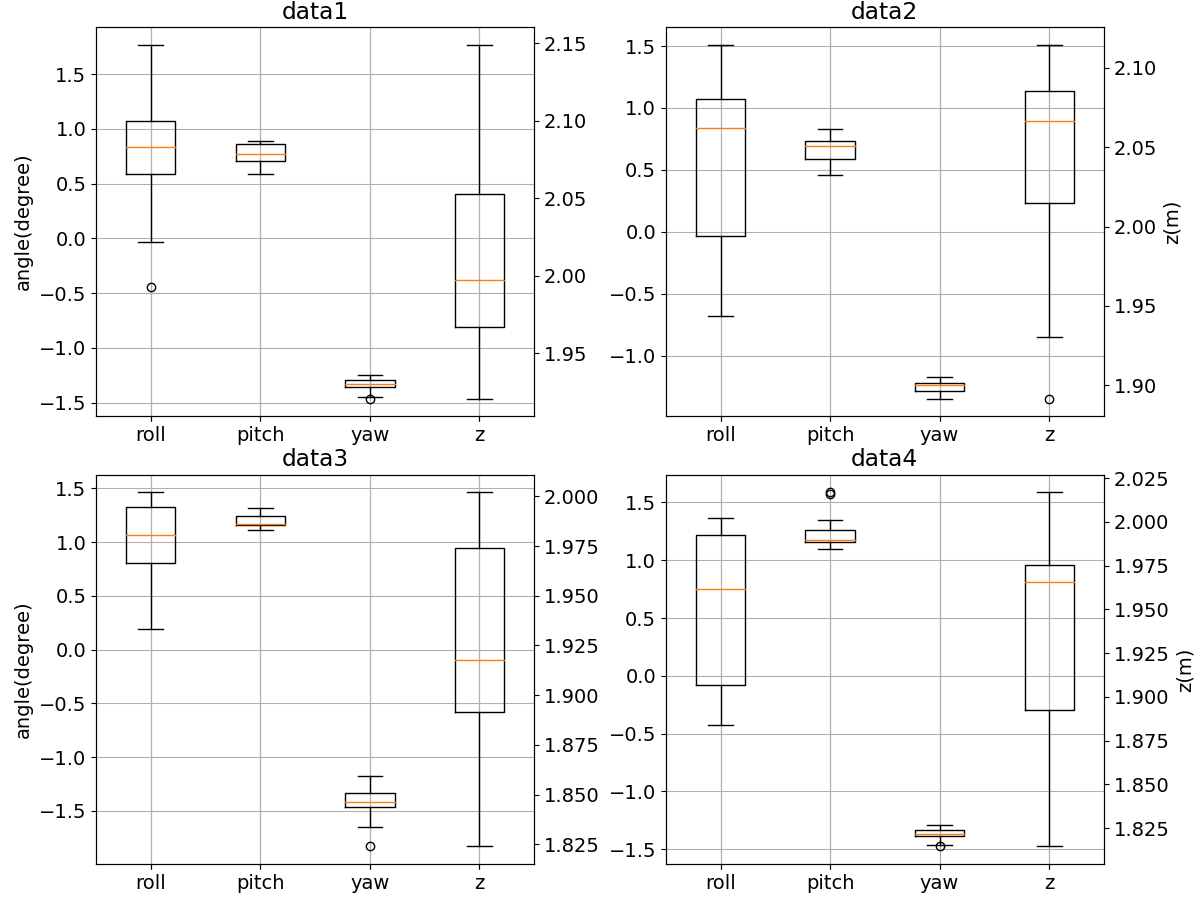}
    \caption{The distribution of LiDAR calibration results in the four data sequences. The fluctuation range of roll, pitch, yaw and z is around 0.5$\degree$, 0.1$\degree$, 0.1$\degree$, 6cm.}
    \label{fig:lidar_box}
\end{figure}

\subsubsection{Calibration results}
We also use the consistency of estimation as an evaluation metric. Each 30-minute data is divided into 1-minute segments. We get an estimated value $\{r_i,p_i,y_i,z_i\}$ from each segment i, and then calculate the standard deviation of all estimates $\{r_i,p_i,y_i,z_i\}_{i=0}^{i=n}$. In order to illustrate the effect of vehicle trajectory on the results, we also calculate the std of selected straight driving segments $\{r_i,p_i,y_i,z_i\}_{i\in S}$. The calibration consistency of the four periods of data is shown in Tab. \ref{tab:lidar_result}. The distribution of the estimates is shown in Fig. \ref{fig:lidar_box}. 

For pitch and yaw angle, the std is around 0.1$\degree$, and the results of all segments don't differ much from those of only straight-line segments. For roll angle, the std is a little large, up to 0.5$\degree$, because of the vehicle tilt in turning time. It decreases to 0.1$\degree$ in only straight driving segments.

\subsection{GNSS/INS Calibration}
\subsubsection{Dataset}
The GNSS/INS data were collected at the same time as LiDAR by Novatel PwrPak7.

\subsubsection{Calibration Results}
\begin{figure}[htp]
    \centering
    \includegraphics[width=\linewidth]{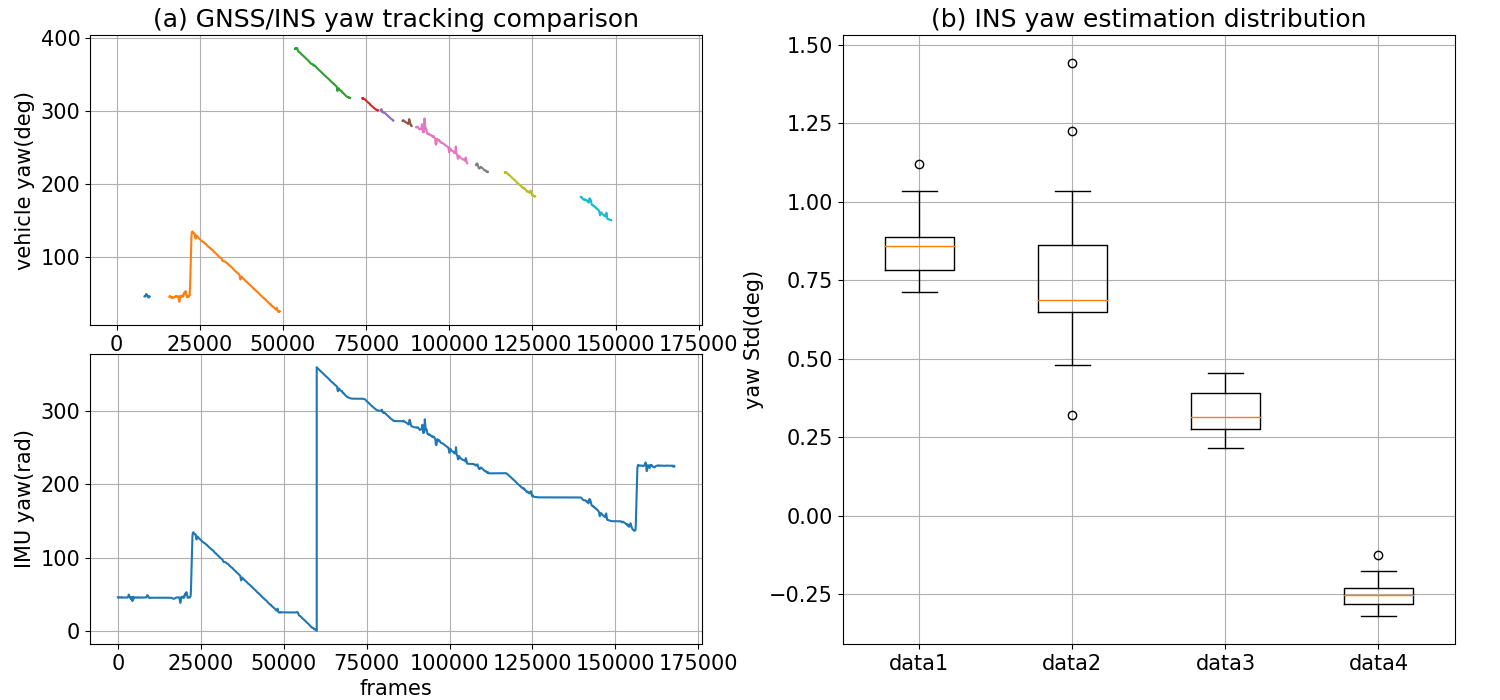}
    \caption{(a) The vehicle yaw angle and IMU yaw angle comparison in data 4. The two curves have the same trend. (b) The distribution of INS calibration results in the four data sequences. The fluctuation range of yaw is around 0.2$\degree$.}
    \label{fig:ins_result}
\end{figure}

As same as LiDAR, each period of data is divided into 1-minute segments and the final estimates of all segments are assembled to calculate the consistency. The standard deviation of estimations is shown in Tab. \ref{tab:lidar_result}. The distribution of the estimations is shown in Fig. \ref{fig:ins_result}(b). The fluctuation range of yaw is around 0.1$\degree$ from all sequences.

To prove the effectiveness of the yaw calibration method, we visualize the vehicle heading direction $\psi_i^v$ and the IMU heading direction $\psi_i^I$ in Eq.\ref{eq:IMU_angle} over an entire data period. The two curves ought to have the same trend and only differ by a fixed offset. Fig. \ref{fig:ins_result}(a) demonstrates the yaw angle tracking result as expected.

\subsection{Radar Calibration}
\subsubsection{Dataset}
The radar data was collected together with LiDAR by Continental ARS 408-21 Long Range 77Ghz Radar. Due to the unsatisfactory quality of the real-world data, we also use simulation data from CARLA \cite{Dosovitskiy17}.

\begin{table}[htp]
    \centering
    \caption{The attribute of radar simulation data in CARLA}
    \begin{tabular}{c|c|c|c}
         \hline
         horizontal fov(deg) & 120 & Max range(m) & 30 \\
         \hline
         vertical fov(deg) & 120 & points per second & 1500 \\
         \hline
    \end{tabular}
    \label{tab:radar_config}
\end{table}

\subsubsection{Calibration}
\begin{figure}[htp]
    \centering
    \includegraphics[width=\linewidth]{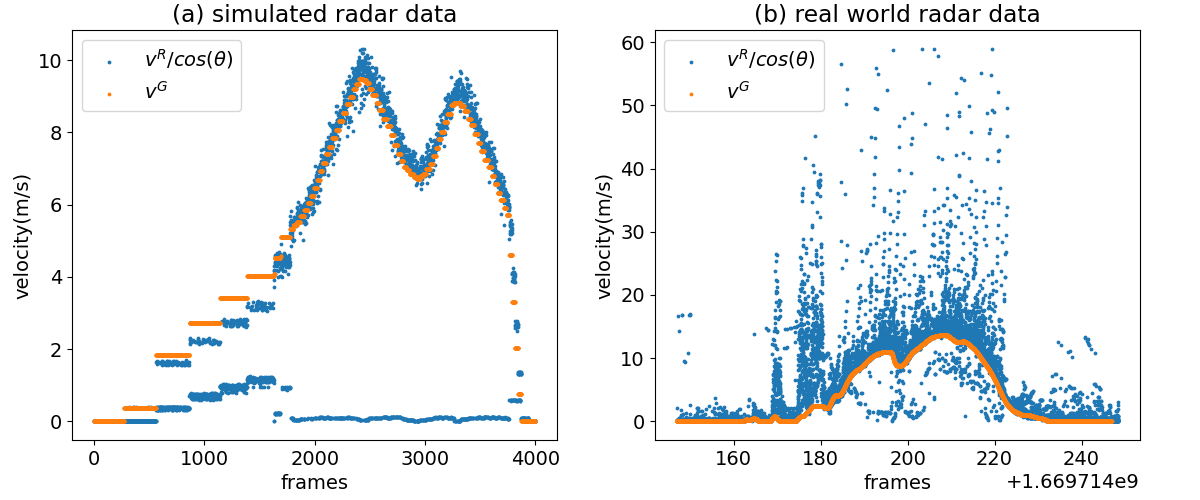}
    \caption{Comparison between $v^G$ and $v^R/cos(\theta)$. (a) For simulation data with yaw = 0, the two angles are close at many points. (b) For real-world(data1), where the yaw offset should be close to zero, there exist many noisy points.}
    \label{fig:radar_noise}
\end{figure}

\begin{figure}[htp]
    \centering
    \includegraphics[width=\linewidth]{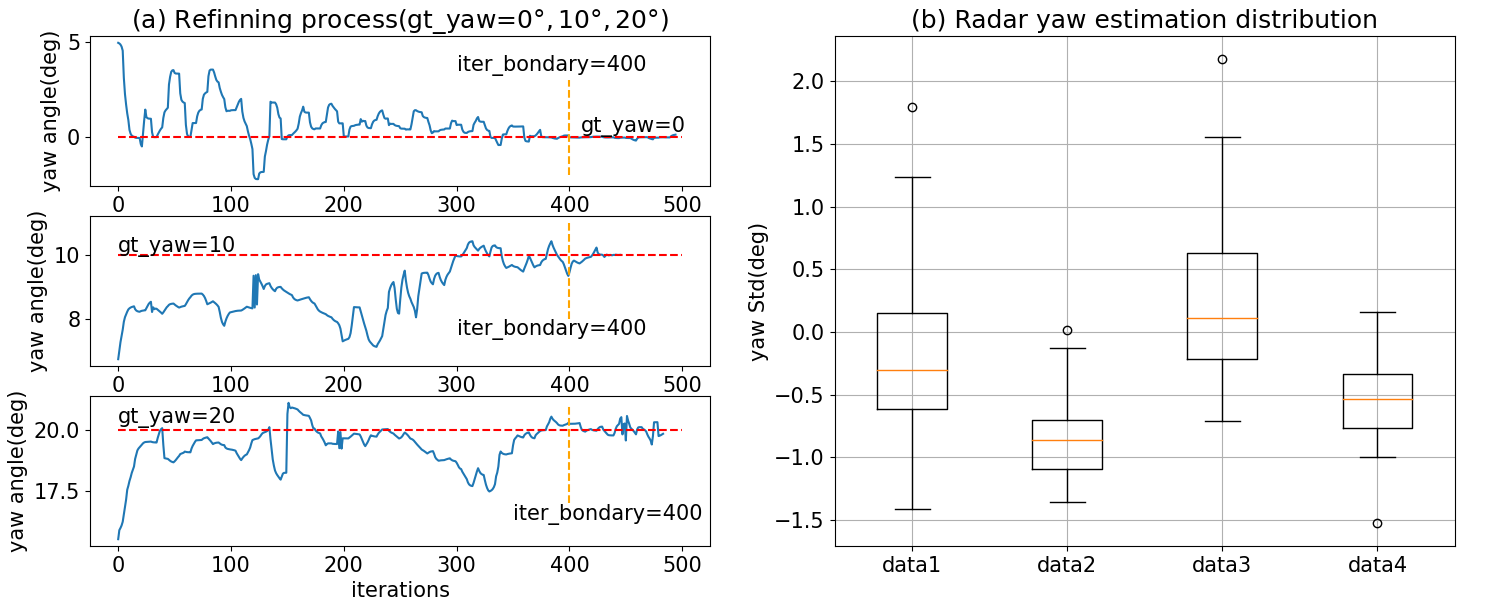}
    \caption{(a) The refined process in velocity method using simulation data. The ground-truth yaw is $0\degree, 10\degree, 20\degree$ and the given initial guess is $5\degree, 5\degree, 15\degree$ (b) The distribution of radar calibration results in the four data sequences. The fluctuation range of yaw angle is around 0.5$\degree$.}
    \label{fig:radar_result}
\end{figure}

We first apply the method only using velocity information. However, we found it performs poor on real-world data. It is because the object velocity in radar measurements has a large noise. As shown in Fig. \ref{fig:radar_noise}, the simulation data basically conforms to Eq.\ref{eq:radar_vel}, but the real-world data has many stray points. So we only apply this method on simulated data. The absolute error is shown in Tab. \ref{tab:radar_result}, smaller than 0.1$\degree$. We also visualize the iteration process in Fig. \ref{fig:radar_result}(a). By giving a 5 degree deviation in the initial guess, the algorithm can converge to the true value after 400 iterations.

\vspace{-0.3cm}

\vspace{-0.3cm}

\begin{table}[ht]
    \centering
    \caption{Radar yaw calibration on simulated data}
    \begin{tabular}{c|c|c|c}
         \hline
          & Duration(s) & Ground truth(deg) & Estimation(deg) \\
         \hline
         1 & 30 & 0 & -0.006993 \\
         \hline
         2 & 30 & 10 & 9.90187 \\
         \hline
         3 & 30 & 20 & 20.0561 \\
         \hline
    \end{tabular}
    \label{tab:radar_result}
    \begin{tablenotes}
    \item initial yaw range $[-A, A] = [-45\degree, 45\degree]$, interval n = 5.
    \end{tablenotes}
\end{table}

For real-world data, we test it with the improved method using the position information. We also use the metric of data consistency. The experimental result of four data period is shown in Tab. \ref{tab:lidar_result}. The distribution of the estimations is shown in Fig. \ref{fig:radar_result}(b). The fluctuation range is about 0.3-0.4$\degree$ in straight-line segments.   

\section{CONCLUSIONS}
We presented SensorX2car, the first open-source calibration toolbox for sensor-to-car extrinsic calibration. It contains online calibration methods for four commonly-used sensors: camera, LiDAR, radar, and GNSS/INS device. The methods can work in common road scenes and have little restrictions on the driving route or road features. The practicality of these methods was verified on simulation and real-world datasets.

In this toolbox, we mainly calibrate the rotation part of the sensor-to-car extrinsic and ignore the translation. In the future, we will extend it to the 6 degrees of freedom parameters for rotation and translation.



\bibliographystyle{IEEEtran}
\bibliography{egbib}

\end{document}